%% file: main.tex
\documentclass[runningheads]{llncs}
\usepackage{makeidx}
\usepackage{graphicx}
\usepackage{sidecap}
\usepackage{amsmath,amssymb} 
\usepackage{color}

\usepackage{float} 
\usepackage{wrapfig}
\usepackage{subcaption}
\usepackage{tabularx}
\usepackage{pbox}
\usepackage{adjustbox}
\usepackage{amsmath}
\usepackage{amssymb}
\usepackage{multirow}
\usepackage{todonotes}

\usepackage{etoolbox}

\begin{document}
	\pagestyle{headings}
	\mainmatter

	\title{Learning to Ignore:\\Fair and Task Independent Representations}

    \author{Linda H. Boedi \and Helmut Grabner}
    \institute{ZHAW School of Engineering,\\Zurich, Switzerland}
    \date{\today}
	
	\maketitle

	\begin{abstract}
    \input{abstract}

	\end{abstract}
	
	\newcolumntype{C}[1]{>{\centering\arraybackslash}p{#1}}

	\section{Introduction}
	\label{sec:introduction}
	\input{introduction}

\section{Learning Invariant Representation}
\input{invariantRepLearning}


\section{Fairness and Interpretability} \label{section:fairness}
\input{fairnessandcausality}


\section{Domain Adaptation}
\input{domainAdaptation} \label{section:domainAdap}

\section{Discussion and Conclusions}
\input{discussion}

	\bibliographystyle{splncs03}
	\bibliography{references}

\end{document}

%% file: abstract.tex
Training fair machine learning models, aiming for their interpretability and solving the problem of domain shift has gained a lot of interest in the last years. There is a vast amount of work addressing these topics, mostly in separation. In this work we show that they can be seen as a common framework of learning invariant representations. The representations should allow to predict the target while at the same time being invariant to sensitive attributes which split the dataset into subgroups. Our approach is based on the simple observation that it is impossible for any learning algorithm to differentiate samples if they have the same feature representation. This is formulated as an additional loss (regularizer) enforcing a common feature representation across subgroups. We apply it to learn fair models and interpret the influence of the sensitive attribute. Furthermore it can be used for domain adaptation, transferring knowledge and learning effectively from very few examples.
In all applications it is essential not only to learn to predict the target, but also to learn what to \emph{ignore}.

%% file: introduction.tex
In June 2020 MIT withdrew Tiny Images\footnote{\url{https://groups.csail.mit.edu/vision/TinyImages/}, 2020/07/10.} a popular vision dataset as researchers found that it is socially biased. Biases in training data are a major issue for machine learning algorithms~\cite{tinyImages}.
Especially, as they are increasingly used to make critical decisions. First, it is important to ensure that those systems are fair and do not discriminate certain groups. Secondly, interpretability of the decisions - "why" the system comes to that conclusion or how important a certain factor for decision making is - are desirable for better understanding. Thirdly, these trained models should generalize well. For many real world situations the data seen during training is different then the data which the models are applied to in production. Domain adaptation tries to transfer knowledge from a source domain (training set) to a particular target domain. In order to cope with these challenges many different approaches have been proposed over the last decade.

\begin{figure}[t]%
\centering
\includegraphics[width=0.8\textwidth]{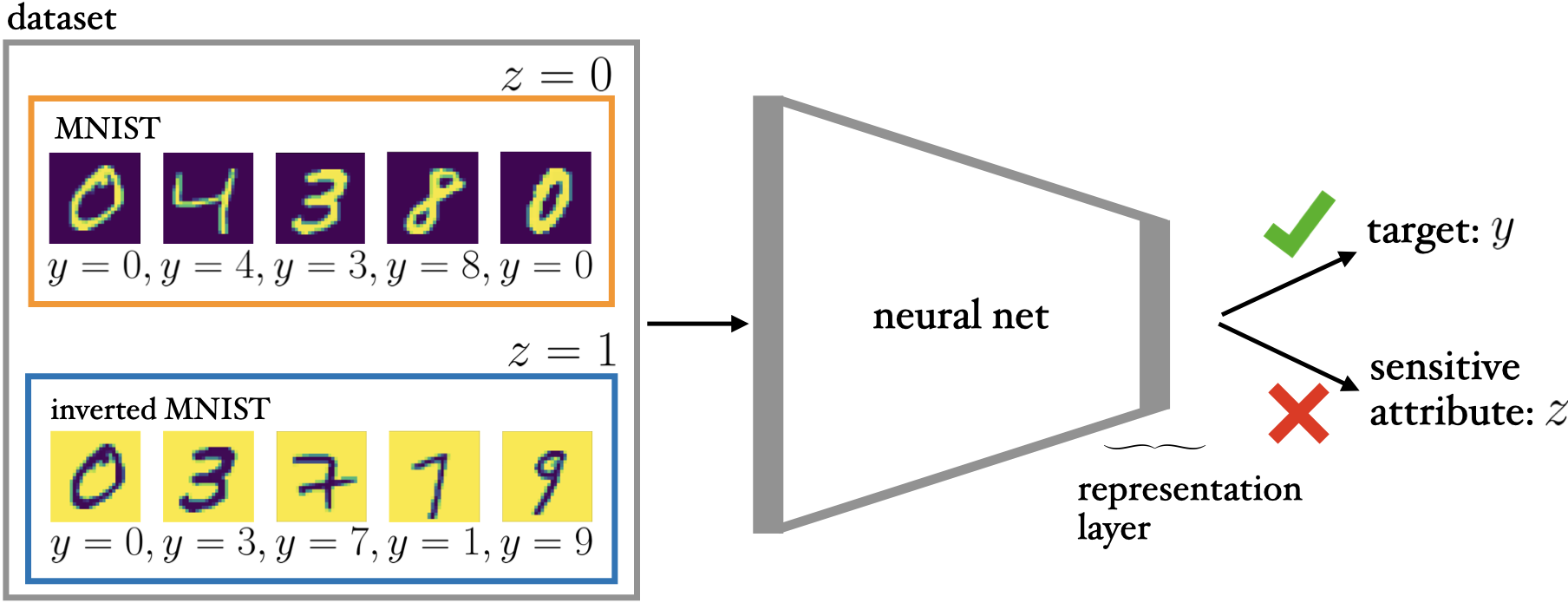}
\caption{Given a set of examples split into groups based on the sensitive attribute $z$, our learnt representation should allow to predict the target variable $y$, but at the same time not be able to predict the sensitive attribute $z$. In the example the digits should be predicted, however the origin (MNIST/inverted MNIST) is irrelevant and should be ignored.}%
\label{fig:teaser}%
\end{figure}

Fairness and domain adaptation seem to be very different topics, but the goal for both is actually learning invariant feature representations.
In this paper we propose a yet simple approach for learning fair representation. A deep learning model is forced to ignore certain information that would allow to draw conclusions about sensitive attributes (fair classifier) or certain areas (domain independent classifier). As seen in Fig.~\ref{fig:teaser} the target variable should be still predictable, but at the same time it should not be distinguishable from which subgroup the examples were taken. The main insight is that similar feature representations for different groups or datasets do not allow us to differentiate between them anymore. To accomplish this, we introduce an affinity loss which is additionally used during the training of a model. Once a fair representation is established, the sensitive attribute can be added back and its impact measured. This paves the way for interpretability or causal reasoning. Furthermore, by reducing the distance between different domains in latent space a more general representation of the dataset is learned which helps to better generalize across domains.

\paragraph{Related Work.}
In order to make machine learning models "fair", works aim at modifying the feature representations of the data~\cite{Zemel2013}, the class label annotations~\cite{ThanhLuong2011} or the data itself~\cite{Quadrianto}.
Learning this latent representation includes an additional cross entropy classification loss~\cite{Beutel}, a decomposition loss~\cite{Quadrianto}, an additional hidden layer for adversarial optimization~\cite{Adel}, distribution matching~\cite{Quadriantoa}, using Variational Autoencoders~\cite{Louizos}, or by learning the representation as an adversarial minimax game~\cite{Xie}. The goal is not only to improve fairness but also to interpret how fairness is enforced.
Such methods build on special network architectures~\cite{Chattopadhyay2019,Louizos2017} or a combination of different machine learning algorithms~\cite{Hartford2016}. 
For the domain adaptation task approaches make use of re-weighting the source samples to better match the target domain~\cite{Huang,Pinheiro2018}, learning shared weights~\cite{Yang} or a common subspace~\cite{Ganin2017}, modifying the network architecture~\cite{Mancini2018,Motiian2017}, or using Generative Adversarial Networks~\cite{Louizos,Ghifary}. Interestingly, if causal aspects are taken into account, predictions can be improved~\cite{Buhlmann2018,Singh2019}.
In the same line, recent work aims for analysing those areas in a common way~\cite{Louizos,Schumann2019,Magliacane2017}.


\paragraph{Contribution.} Our main aim is not to beat any particular method for fairness or domain adaption, it is rather to highlight the commonalities. From a technical side, the most related work probably is the one by Ganin et al.~\cite{Ganin2017} based on Zemel et al.~\cite{Zemel2013}. However instead of adding a new gradient reversal layer, we simply reduce an additional affinity loss during training. Hence, our method can be easily applied to any existing network architecture including classification, regression tasks or auto-encoders\footnote{In this paper we focus on the classification tasks with one categorical sensitive variable.}. Experiments demonstrate that a pretrained network with fixed weights can be simply debiased by adding a fair representation layer. While many approaches struggle with unbalanced datasets both in terms of the target and the sensitive attribute~\cite{Beutel}, our approach is not very negatively impacted. We are able to improve fairness, interpretability and domain adaptation within one very simple approach.


%% file: invariantRepLearning.tex
\paragraph{Problem formulation.} Let $X$ be the entire data set. Each $x \in X$ is an example represented by $m$ attributes and $y \in Y$ its corresponding target variable. Furthermore, let $z$  be the sensitive attribute. We aim to learn a classifier $f_y: x \rightarrow y$ to predict the target variable from the attributes, but at the same time being \emph{unable} to predict the sensitive attribute $f_z: x \rightarrow z$.

In order to achieve this we build a (low dim.) representation $g(x)$ which allows for predicting the target $y$ but \emph{not} the sensitive attribute $z$. Our aim is to make the representation as similar as possible with respect to the sensitive attribute. Hence, being invariant features. Many modern deep learning architectures can be seen as having such a representation layer, having the advantage that any pre-trained model can be used and later fine-tuned. 

\begin{SCfigure}[1][bt]%
\centering
\caption{Example architecture of a neural network with one ore more hidden layers $h$ (indicated with the dotted lines). The last hidden layer ($r_1,..., r_n$), is reffed as representation and used for calculated our proposed affinity loss.}%
\includegraphics[width=0.55\textwidth]{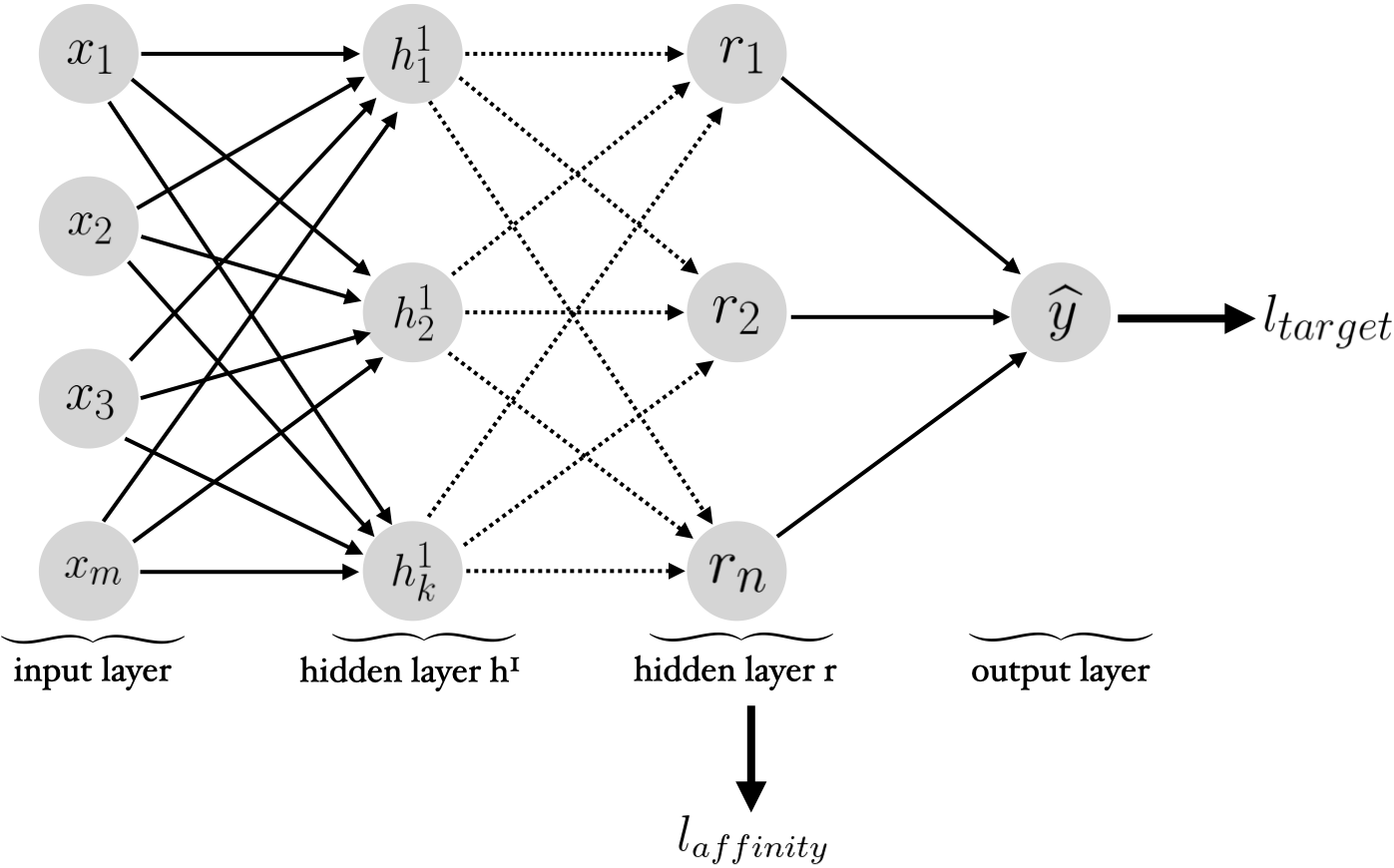}
\label{fig:modelarchitecture}%
\end{SCfigure}

In the setting of training a neural network, typically a loss $l_{target}$ (e.g., cross entropy) is minimized in order to predict the target. We propose to add another loss term $l_{affinity}$ which serves as regularizer. See a visualization in Fig.~\ref{fig:modelarchitecture}. The neural network is then trained on the combined loss
\begin{equation}
l_{total} = l_{target} + \lambda \cdot l_{affinity}.
\end{equation}
If the weight $\lambda$ of the affinity loss is set to zero, the model is trained normally without the new loss. If $\lambda$ is very large the neural network optimizes on the affinity loss, ignoring the target loss. The fairness of a model increases with $\lambda$ but might result in lower accuracy.

In the following we derive $l_{affinity}$. The sensitive attribute splits the dataset in one or more subgroups. For simplicity we focus on two subgroups in the following. Considering the two sets $X_1, X_2$ split by the sensitive attribute $z$. To be unable to distinguish between these two sets the following must hold
\begin{equation}
\forall x_1 \in X_1 \ \exists x_2 \in X_2 : g(x_1) = g(x_2).
\end{equation}
In other words, for each sample there must be at least another sample with a different sensitive attribute having the same representation. Technically the loss is minimizing the closest distance of them.

The learnt representation should still allow to predict the target $y$. Trivial representations are avoided by the combination of the loss term (see above). However, our experiments show that it is beneficial to add a more strict constraint so that the two examples $x_1$ and $x_2$ are from the same class, i.e., $y_1 = y_2$. This avoids mixing up classes and yields significantly better performance. Averaging over all examples and all classes gives
\begin{equation}
    l_{affinity} = \frac{1}{|Y| |X_1|}\sum_{y \in Y} \sum_{x_1 \in (X_1 | y_1 = y)} \min_{x_2 \in (X_2 | y_2 = y)} d(g(x_1), g(x_2)),
    \label{eq:1a}
\end{equation}
where $d(\cdot, \cdot)$ is an arbitrary distance function.

\paragraph{Implementation Details.} In order to implement Eq.~\eqref{eq:1a} we use a nearest neighbor with the $L_1$ norm.
To speed the training up we do not use the whole dataset, but only calculate the affinity loss on the mini-batches. 

\subsection{Experiments\label{sec:initialExp}}
We show the basic behavior of our method on an illustrative experiment based on the well known \emph{MNIST} dataset of handwritten digits\footnote{\url{http://yann.lecun.com/exdb/mnist/}, 2020/07/10.}. 
Additionally we created the \emph{MNIST-I}, which contains all original \emph{MNIST} images, but inverted. Together it forms our dataset where the sensitive attribute indicates if the digit originates from the \emph{MNIST-I} or the original \emph{MNIST}. As target we still want to predict which number is depicted. 

We train a simple neural network with two 128- and a single 20-width ReLU hidden layer as representation layer. For training, a batch size of 128 samples is used and the weight of the proposed affinity loss is set to $\lambda = 0.01$. 

\paragraph{Embedding.}
In order to analyze our learnt representation, we perform a t-Distributed Stochastic Neighbor Embedding (t-SNE) on the representation layer. It models the higher-dim. data by a low-dim. point such that similar objects lay closer together and dissimilar ones further away. 

\begin{SCfigure}
     \begin{tabular}{m{1.7cm}C{3cm}C{3cm}}
      &{baseline}& {our approach} \\
    {target (digits)}&\parbox{3cm}{\includegraphics[width=.25\textwidth]{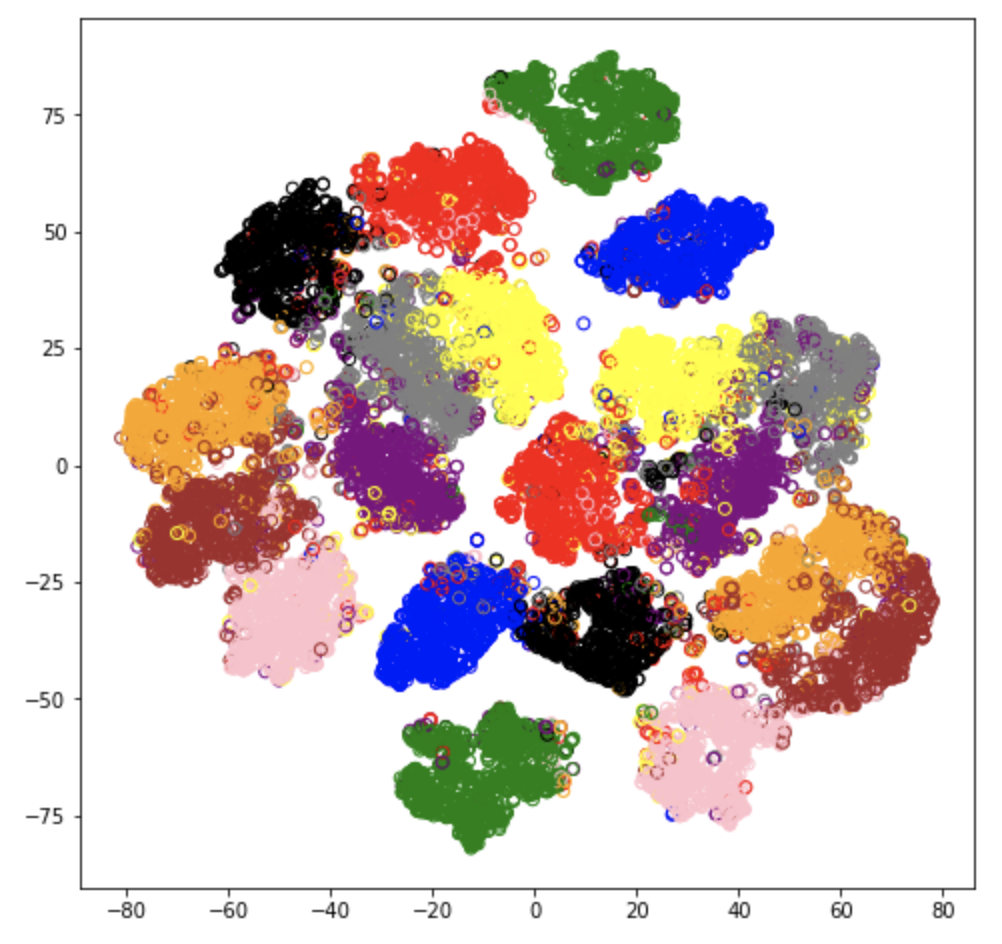}} & \parbox{3cm}{\includegraphics[width=.25\textwidth]{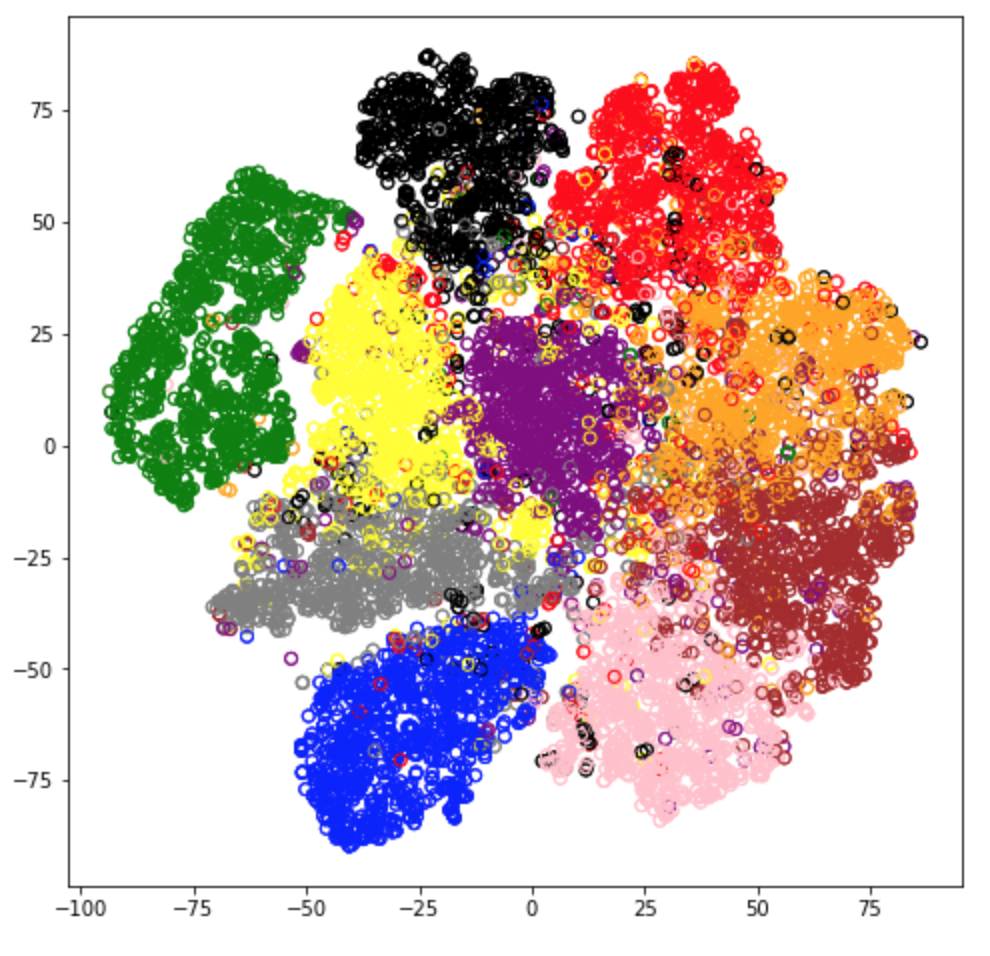}} \\ 
    {sensitive attribute (MNIST/
    MNIST-I)}&\parbox{3cm}{\includegraphics[width=.25\textwidth]{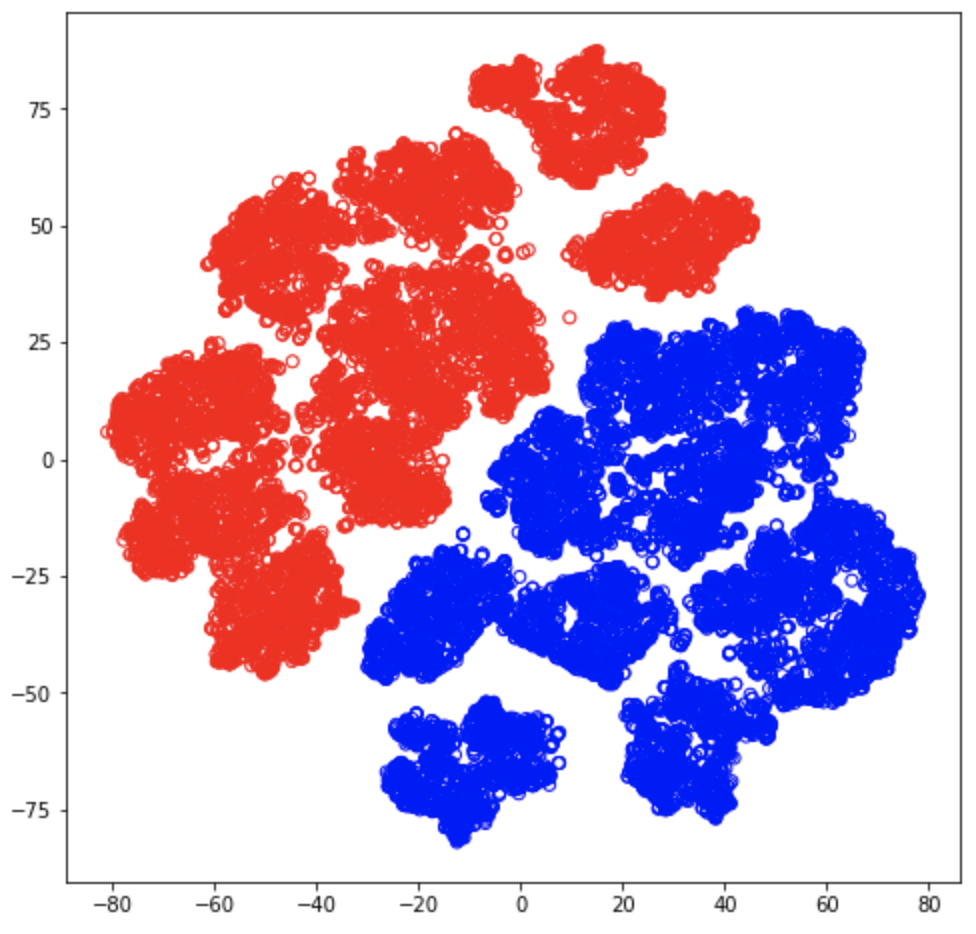}} &
    \parbox{3cm}{\includegraphics[width=.25\textwidth]{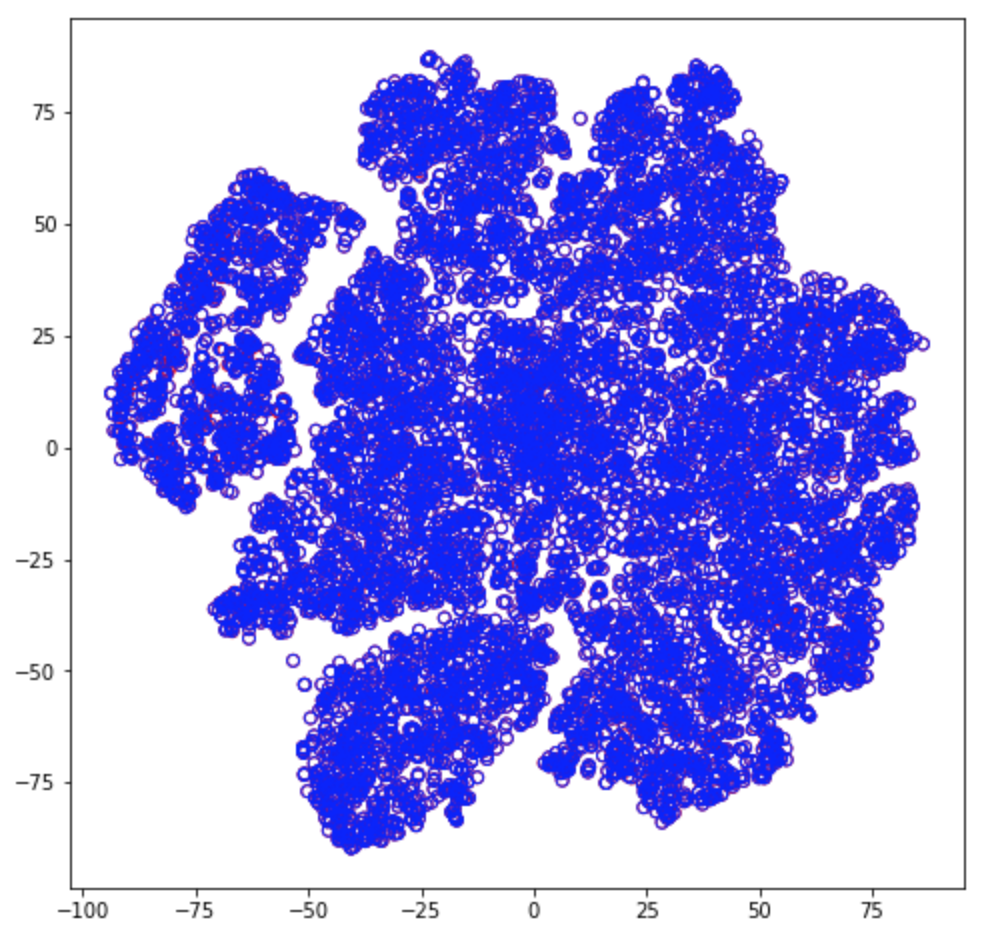}}\\
    \end{tabular}
    \caption{2-dim. t-SNE plot showing the learnt feature representations colored based on digit (top row) and on the sensitive attribute (bottom row). By using our approach, the representation shows distinguishable digit clusters but the dataset origin cannot be traced. Best viewed in colour!}
    \label{tab:t-SNE-MNIST}
\end{SCfigure}%

Fig.~\ref{tab:t-SNE-MNIST} depicts the comparison of two models, trained without (baseline) and with the affinity loss. The baseline model learns two clusters for each digit (one normal and one inverted) and the groups can easily be separated by the sensitive attribute (MNIST/MNIST-I). In contrast, adding the proposed affinity loss into the training process shows that the two groups are highly overlapping, i.e., not being distinguishable anymore, creating a fair (more in Sec.~\ref{section:fairness}) and more general (only one cluster per digit) feature representation. The digits can still be predicted very accurately as their clusters are kept very distinct from each other. The effect of generality is also used for domain adaptation in Sec.~\ref{section:domainAdap}.

\paragraph{Predictions}
After training, we fix all hidden layers and do a normal retrain of the output layer. The output layer is not only trained to predict the digits but also if the sample comes from the original MNIST or inverted MNIST-I dataset. The model trained with our approach should struggle in learning the origin of the sample (inverted/not inverted). In fact, our fair model predicts the target class with 93\% accuracy (4\% higher than the baseline), whereas the sensitive attribution is hardly predictable anymore (around 57\% accuracy). The baseline model can easily predict (nearly 100\%) the sensitive attribute.


\paragraph{Hyperparameter.}
If the weight $\lambda$ of the affinity loss is set to zero, we train the model only based on the target loss not focusing on making the model fair (see Tab.~\ref{tab:MNIST-MNISTI-Weight}, trained 5 times and averaged). For $\lambda$ equals 0.01 we get the fairest model, as the numbers are predicted well (even better than the baseline model does) and the origin dataset of the input samples can hardly be predicted. For a $\lambda$ of 0.1 the model gets as fair as possible by not learning anything at all. 
If the representation layer is chosen too small (smaller than 5 nodes) the model does not perform well in predicting the digits accurately (see Tab.~\ref{tab:MNIST-MNISTI-Layers}). If the number of nodes is getting too large (more than 50 in this example) the model gets unfair again suffering from the curse of dimensionality.

\begin{table}[tb]
  \centering
      \caption{Influence of Hyper-parameters on fairness and accuracy.}
      \begin{subtable}{\textwidth}
      \caption{Tuning $\lambda$ balances accuracy vs. fairness.}
    \begin{tabular}{|l|C{2cm}C{2cm}C{2cm}C{2cm}C{2cm}|}
    \hline
    \textbf{$\lambda$} & 0 & 0.0001 & 0.001 & 0.01 & 0.1 \\\hline\hline
    \textbf{accuracy} & 0.89 & 0.9 & 0.93 & 0.93 & 0.1 \\
    \textbf{digits} & $\pm$ 0.02 & $\pm$ 0.02 & $\pm$ 0.01 & $\pm$ 0.01 & $\pm$ 0.005\\\hline
    \textbf{accuracy} & 0.995 & 0.99  & 0.89 & 0.57 & 0.5\\
    \textbf{sensitive} & $\pm$ 0.005 & $\pm$ 0.005 & $\pm$ 0.02 & $\pm$ 0.03 & $\pm$ 0.005  \\\hline
    \end{tabular}
    \label{tab:MNIST-MNISTI-Weight}
    \end{subtable}
    \begin{subtable}{\textwidth}
    \caption{Too small representations (obviously) do not allow for good classification results. A too large embedding space suffers from the curse of dimensionality.}
    \begin{tabular}{|l|C{1.6cm}C{1.6cm}C{1.6cm}C{1.6cm}C{1.6cm}C{1.6cm}|}
    \hline
    \textbf{layer size} & 1 & 5 & 10 & 20 & 50 & 100 \\\hline\hline
    \textbf{accuracy} & 0.1  & 0.78  & 0.93  & 0.93 & 0.95 & 0.96 \\
    \textbf{digits} & $\pm$ 0.05 & $\pm$ 0.03 & $\pm$ 0.01 & $\pm$ 0.01 & $\pm$ 0.03 & $\pm$ 0.04\\\hline
    \textbf{accuracy} & 0.5 & 0.54  & 0.57 & 0.57 & 0.74 & 0.85\\
    \textbf{sensitive} & $\pm$ 0.05 & $\pm$ 0.02 & $\pm$ 0.02 & $\pm$ 0.03 & $\pm$ 0.04 & $\pm$ 0.05\\\hline
    \end{tabular}
    \label{tab:MNIST-MNISTI-Layers}
    \end{subtable}
\end{table}%

%% file: fairnessandcausality.tex
Simply removing the sensitive attributes from a dataset is insufficient for eliminating their biases as there almost always exists an indirect influence of the sensitive information~\cite{Pedreschi2008}. 
Our approach learns a feature representation of the data preserving general information but enforcing not to learn sensitive characteristic information. 

\begin{SCfigure}[1][bt]%
   \centering
\includegraphics[width=0.55\textwidth]{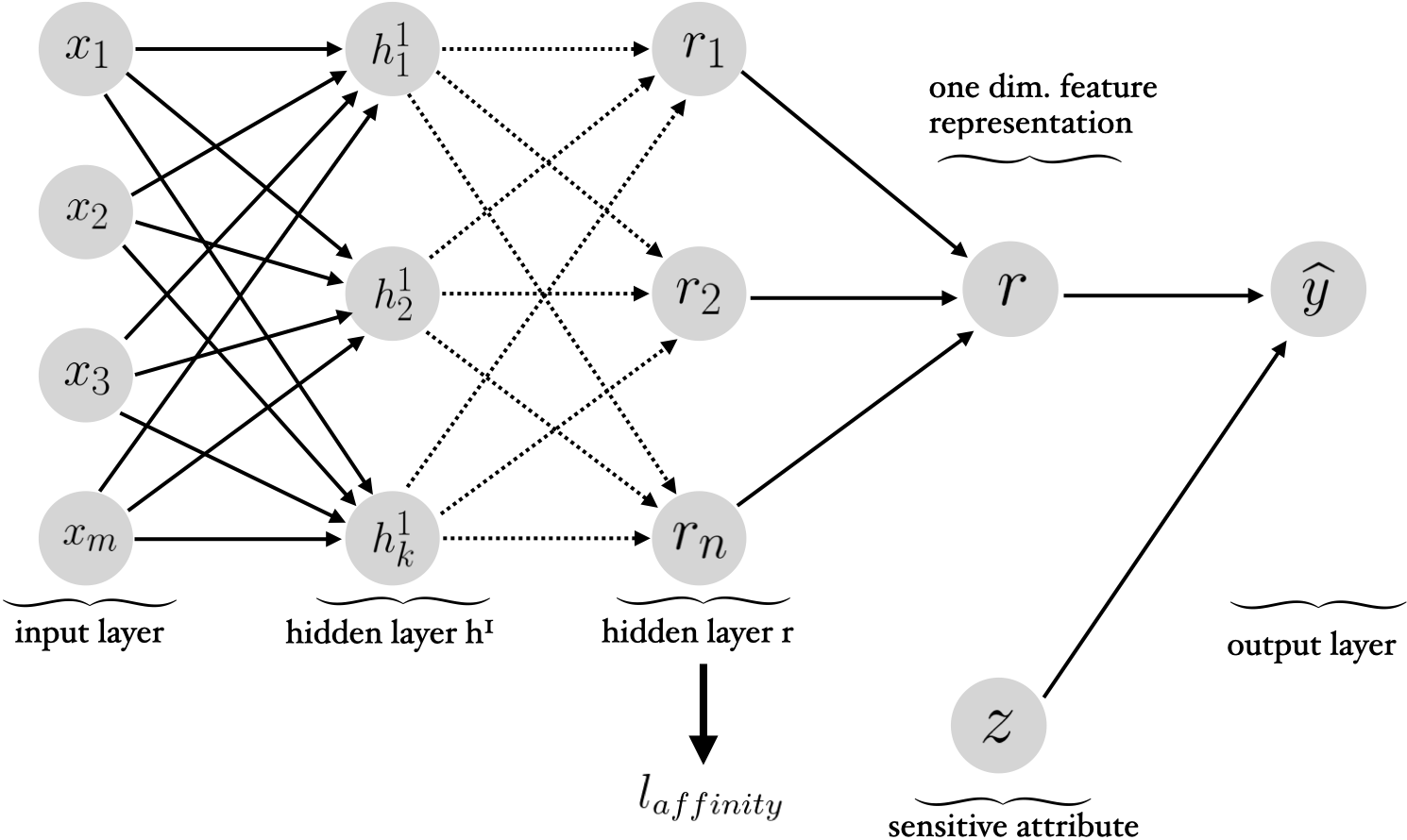}
\caption{A simple linear unit is added to the fair model representation. Furthermore the sensitive attribute $z$ is reattached. As the both units are uncorrelated, we use the weights as interpretation for the importance of the sensitive attribute for the final classification.}
   \label{fig:CausalityArchitect}
\end{SCfigure}

After the fair training of the model we are able to interpret the classification and investigate in the influence of the sensitive attribute on the classification task~\cite{Peters}. We do so by reattaching the sensitive attribute $z$ to the fair model again, see Fig.~\ref{fig:CausalityArchitect}. For better interpretability the fair feature representation is is linearly combined, forming $r$. The reattachment of $z$ and its interpretation is possible as $r$ is trained to be independent of $z$ (see also~\cite{Amsterdam2019})
\begin{equation}
\hat{y} = f(w_r r + w_z z + b), \text{with } z \perp \!\!\! \perp r
\end{equation}
where $w_r$ and $w_z$ are the learnt weights of the neural network, b the bias term and $z$ the sensitive attribute and $f(\cdot)$ the transfer function (e.g., linear or sigmoid). The weights $w_r$ and $w_z$ of $r$ and $z$, respectively, indicate how large the influence of the sensitive attribute on the classification is~\cite{Peters}. 
In the following experiments a model trained without the affinity loss using the same architecture as the fair one is reffed as baseline.

\subsection{Fairness Measures}\label{subsection:measures}
There are a lot of different fairness measures used for classification~\cite{Verma2018,Beutel,HardtGoogle,Zafar2016}. Two commonly used ones are summarized in the following. Let $\hat{y}$ be the output of the classifier, $y$ the true label and $z$ the sensitive attribute.

\paragraph{Equality of Opportunity/ Equality Gap.} The most common measure is the so-called equality of opportunity. It is reached if the groups $z_1$ and $z_2$ defined by the sensitive characteristic have equal true positive rates (TPR), i.e., $TPR_{z=z_1} = TPR_{z=z_1}$. 
The equality gap is then calculated as 
\begin{equation}
{P}(\hat{y} = 1 |  z = z_1, y = 1) - {P}(\hat{y} = 1|  z = z_2, y = 1) = | TPR_{z=z_1} - TPR_{z=z_2} |.
\end{equation}

\paragraph{Parity Gap.} The parity gap is calculated as independence between prediction $\hat{y}$ and sensitive attribute $z$ for positive predictions, i.e. 
\begin{equation}
|P(\hat{y} = 1|Z=z_1) - P(\hat{y} = 1|Z=z_2) |    
\end{equation}
For binary case of the sensitive attribute, in medical settings, it is the same as the average treatment effect (ATE)~\cite{Amsterdam2019}.

It is important to note that it is difficult to minimize all fairness metrics at the same time. The appropriate metric depends on the application, but most often the equality of opportunity is targeted. Be aware, that there are some trivial models which yield good results (very small gaps) such as models with very low TPR. For some tasks, the compromise may not even be possible, such as predicting whether someone can give birth. There is a clear causal relationship to the gender; thus, if this information (including implicit information) is removed, it becomes impossible for any classifier to make a correct prediction.

\subsection{Experiment: Adult Dataset} \label{sec:adultExperiment}
The popular \emph{Adult} income dataset\footnote{\url{https://archive.ics.uci.edu/ml/datasets/adult}, 2020/07/10.} from the UCI is used for further experiments and the results are compared with other papers. The task is to predict whether or not an individual is earning more than \$50K per year. The samples are annotated with 14 different attributes from gender and educational level to number of work hours per week. The gender attribute is used as binary sensitive attribute for the affinity loss during training. The dataset is split into 26’049 samples for training, 6,512 samples for validating and 16,281 for testing.

We train our model with a batch size of 512 samples using a network with one hidden layer of 128 and another one with 20 nodes on whose feature representation the affinity loss is calculated. We compare our model with several state-of-the-art approaches 
as well as against the baseline. Results are summarized in Tab.~\ref{tab:fairnessAdult}. Our approach achieves a similar fairness level compared to other approaches. Consistently, our feature representation promoted fairness criteria with only a small penalty in accuracy even though the dataset is heavily skewed.

\begin{table}[tb]
\centering
  \caption{Our approach compared with a baseline and different other approaches concerning accuracy and fairness on the Adult dataset.}
  \begin{tabular}{|l|C{2cm}|C{2cm}C{2cm}|C{3cm}|}
    \hline
    \textbf{approach} & \textbf{accuracy} & \textbf{parity gap} & \textbf{equ. gap} & \textbf{equ. gap (TNR)}\\\hline\hline 
    {baseline}& 0.85 &  0.18 & 0.088& 0.072\\\hline 
    {ours} & 0.82& 0.065& 0.02& 0.015 \\\hline
    {Quadrianto'18~\cite{Quadriantoa}} &0.81& - & 0.04 & -\\
    {Adel'19~\cite{Adel}} & 0.89& 0.13& - & -\\
    {Zemel'13~\cite{Zemel2013}} &0.82 & - & - & -\\
    {Quadrianto'17~\cite{Quadrianto}} & 0.84 & - &0.017& -\\
    {Beutel'17~\cite{Beutel}} & 0.82 &0.12 & 0.07& 0.04\\
    {Louizos'17~\cite{Louizos}} & 0.82 & - & 0.05 & -\\
    {Xie'17~\cite{Xie}} & 0.84 & - & - & - \\\hline 
    \end{tabular}
    \label{tab:fairnessAdult}
\end{table}%

\paragraph{Fair Representation.}  We train two baseline models once inputting all the features including the sensitive attribute and once removing this attribute. We assume that simply removing the sensitive attribute does not help to omit the gender bias~\cite{Pedreschi2008}. The performances are compared with our fair model and if the sensitive attribute is added back, see Fig.~\ref{fig:CausalityArchitect}. For each model we retrain the last layer (hidden layers are fixed) to predict once the gender of the input samples and once the income. Results are summarized in Tab.~\ref{tab:CausalityGender}. The accuracies of the baseline models lay very close together, which shows that information about the gender attribute is indeed still hidden in the input data. Our model is not able to classify the genders correctly just labeling almost all samples as male. The performance of our model with reattached gender attribute is similar to the baseline models. This shows that the sensitive attribute helps the model to perform better. 

\begin{figure}[tb]%
\centering
\begin{center}
\begin{subfigure}[t]{0.4\textwidth}
\includegraphics[width=1\textwidth]{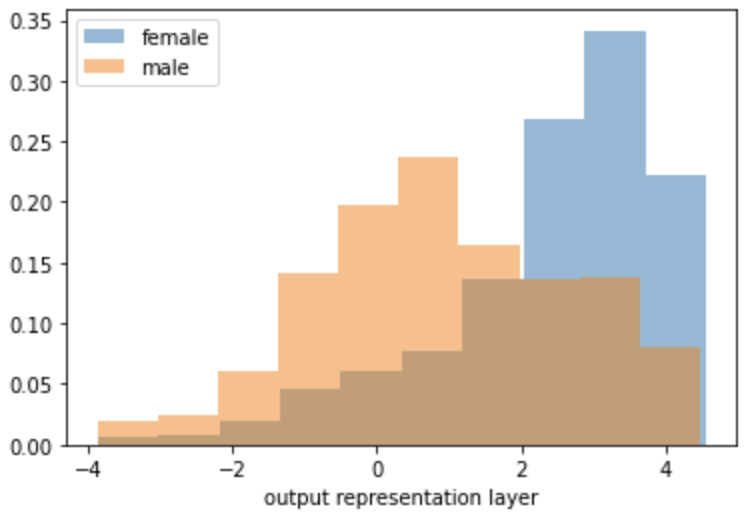}
\subcaption{baseline}
\end{subfigure}
\begin{subfigure}[t]{0.4\textwidth}
\includegraphics[width=1\textwidth]{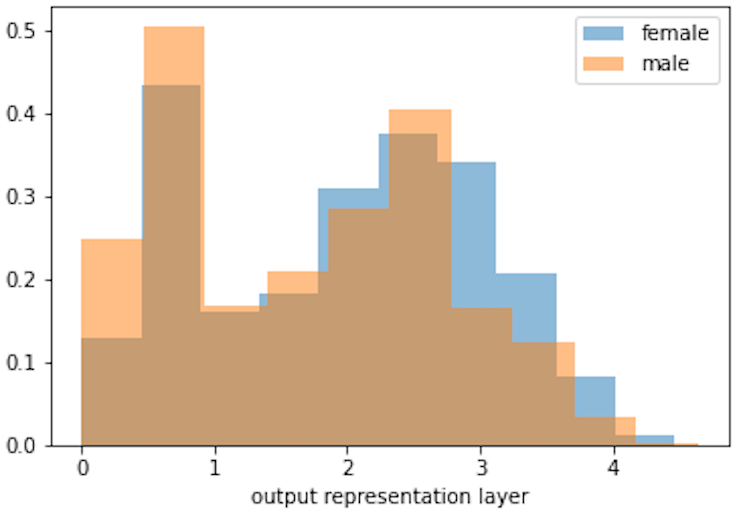}
\subcaption{our approach}
\end{subfigure}
\end{center}
\vspace{-0.6cm}
\caption{Histogram of the outputs of $r$ (see Fig.~\ref{fig:CausalityArchitect}) for male and female samples. Without enforcing any fair training the learnt distributions carry a lot of information about the sensitive attribute (a) in contrast to our approach with overlapping distributions (b), not allowing to trace back the sensitive attribute.}%
\label{fig:CausalityHists}%
\end{figure}

The histograms of the fair one-dim. representation (see Fig.~\ref{fig:CausalityArchitect}) for the male and female samples in Fig.~\ref{fig:CausalityHists} show a very similar distribution. Hence, this supports the assumption that our model contains a fair representation of the data.

\begin{table}[t]
  \centering
      \caption{Accuracy and fairness measures predicting income and gender attribute ($z$) on the Adult dataset with different training approaches.}
      \begin{tabular}{|l|C{2cm}C{2cm}|C{2.5cm}C{2.3cm}|}
            \hline
    \multicolumn{1}{|c|}{} & \multicolumn{2}{c|}{\textbf{baseline}}& \multicolumn{2}{c|}{\textbf{our approach}} \\\hline
    & 
    {with z}& 
    {without z} &
    {} &
    {z reattached} \\ \hline\hline
    
    \textbf{accuracy income} & 0.85  & 0.85  & 0.81 & 0.83 \\\hline
    \textbf{parity gap} & 0.16 & 0.17 & 0.05 & 0.15 \\
    \textbf{equality gap} & 0.069  & 0.074 & 0.0075 & 0.1\\\hline
    \textbf{accuracy female} &  0.46 & 0.25  & 0.027  &1.0\\
    \textbf{accuracy male} & 0.84 & 0.86 & 0.98  & 1.0\\
    \hline
    \end{tabular}
    \label{tab:CausalityGender}
\end{table}%

\paragraph{Interpretablity of the sensitive attribute.}\label{par:interpretability} The influence of the gender attribute on the classification of the input samples concerning the income is evaluated. The weight of the fair one-dim. feature representation $r$ and the weight of the input of the sensitive attribute $z$ are compared (see Fig.~\ref{fig:CausalityArchitect}). We consider the samples where the income is larger than \$50K. The weight for $z$ lays around 4, whereas the weight for $r$ is approximately 1, thus much smaller. This shows that the input of the sensitive attribute has indeed a large influence on the classification as it holds highly valuable information.

\subsection{Experiment: CelebA}
The \emph{CelebA} image dataset\footnote{\url{http://mmlab~.ie.cuhk.edu.hk/projects/CelebA.html}, 2020/07/10.} is significantly more complex than the \emph{MNIST} or \emph{Adult} dataset. This record contains a total of 202,599 images of celebrities, each with 40 attributes. 162,770 images are used for training, 19,867 for validating and the rest for testing. The annotated attributes reflect appearance of the celebrities as well as the emotional state (e.g. smiling), gender, attractiveness and age. The gender attribute is used as a binary sensitive characteristic and attractiveness as a target label for the classification of the images.

As model we use a fixed VGG19 net trained on imagenet (to speed up the training process and reduce complexity) and an additional hidden layer with 124 nodes. 
Tab.~\ref{tab:fairnessCelebA} compares results with different weights $\lambda$ and shows that we can in fact debias the pretrained VGG net. The \emph{CelebA} dataset is heavily skewed; around $\sim{77\%}$ of the images showing women are labeled as attractive, compared to $\sim{23\%}$ of men. If $\lambda$ is strong enough, the influence of the skew on the fairness disappears. The downside is the decrease of accuracy to only $\sim{55\%}$ as the TNRs for female and male are getting low. Please note, the comparison with Quadrianto~\cite{Quadrianto} is not too accurate as our baseline already has a lower accuracy.

\begin{table}[tb]
  \centering
  \caption{Fairness of different models on the CelebA dataset. A lower weight of $\lambda$ keeps the accuracy higher, but improves fairness only to a limited amount. A higher weight decreases the accuracy significantly but makes the model fair.}
    \begin{tabular}{|l|C{1.5cm}|C{1.5cm}C{1.5cm}C{1.5cm}|C{1.5cm}C{1.5cm}|}
    \hline
    \multicolumn{1}{|c|}{} & \multicolumn{1}{c|}{\textbf{baseline}}& \multicolumn{3}{c|}{\textbf{our approach}}& \multicolumn{2}{c|}{Quadrianto'18\cite{Quadrianto}} \\
    & 
    \textbf{}& 
    \textbf{$\lambda=0.05$} &
    \textbf{$\lambda=0.07$} &
    \textbf{$\lambda=0.1$} &
    \textbf{fair} &
    \textbf{baseline}\\\hline\hline
    \textbf{accuracy} & 0.73 & 0.64 & 0.62& 0.55 & 0.8 & 0.8\\\hline
    \textbf{parity gap} & 0.47  & 0.23 & 0.22 & 0.0038 & - & - \\
    \textbf{equality gap} & 0.31  & 0.25 & 0.23 & 0.018 & 0.19 & 0.34\\
    \hline
    \end{tabular}
  %
    \label{tab:fairnessCelebA}
\end{table}%

\begin{table}[!tb]
  \centering
      \caption{Accuracies and fairness measures predicting gender attribute $z$ and income on the CelebA dataset with different training approaches.}
       \begin{tabular}{|l|C{2cm}C{2cm}|C{2.5cm}C{2.3cm}|}
      \hline
    \multicolumn{1}{|c|}{} & \multicolumn{2}{c|}{\textbf{baseline}}& \multicolumn{2}{c|}{\textbf{our approach}} \\\hline
    & 
    {with z}& 
    {without z} &
    {} &
    {z reattached} \\ \hline\hline
    
    \textbf{accuracy} & 0.73  & 0.74  & 0.62 & 0.7\\\hline
    \textbf{parity gap} & 0.47   & 0.49   & 0.22 & 0.63\\
    \textbf{equality gap} & 0.31   & 0.32  & 0.23  & 0.59\\
    \hline
    \textbf{accuracy female} & 0.86 & 0.84  & 0.88 &1.0\\
    \textbf{accuracy male} & 0.83 & 0.85   & 0.25 & 1.0\\
    \hline
    \end{tabular}
    \label{tab:CausalityGenderCelebA}
\end{table}%
\begin{figure}[!tb]
\centering
\begin{center}
\begin{subfigure}[t]{0.4\textwidth}
\includegraphics[width=1\textwidth]{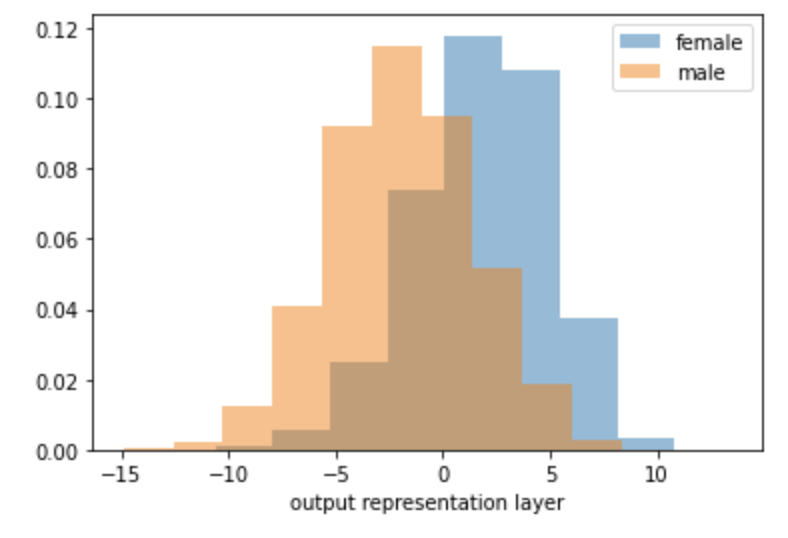}
\subcaption{baseline}
\end{subfigure}
\begin{subfigure}[t]{0.4\textwidth}
\includegraphics[width=1\textwidth]{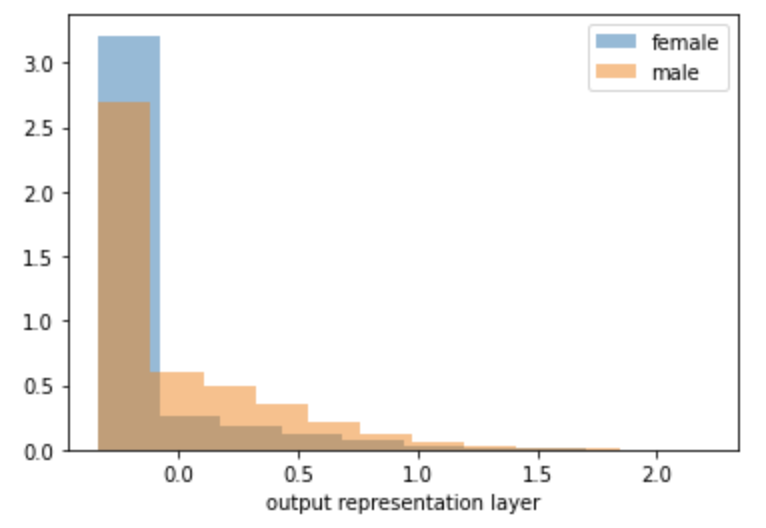}
\subcaption{our approach}
\end{subfigure}
\end{center}
\vspace{-0.6cm}
\caption{Histogram of the outputs of $r$ for male and female samples for \emph{CelebA}.}%
\label{fig:CausalityHistsCelebA}%
\end{figure}

\paragraph{Fair representation \& Intepretability.} The histograms of the the fair one-dim. representation (see Fig.~\ref{fig:CausalityArchitect}) for the male and female samples in Fig.~\ref{fig:CausalityHistsCelebA} show a very similar distribution, supporting the assumption that our model contains a fair representation of the data. The influence of the gender attribute on the classification, see Tab.~\ref{tab:CausalityGenderCelebA}, is checked with the same approach as described in Sec.~\ref{sec:adultExperiment}. The similar accuracies of the baseline models show that information about the gender attribute is indeed still hidden in the input data. The reattached sensitive attribute helps the fair model to perform better in classifying faces as attractive. This can also be seen in the weight of the sensitive attribute $z$ with around 1.5, compared to the one of the fair one-dim. feature representation $r$ with around 0.7.

%% file: domainAdaptation.tex
Data used for training a model might not be the same as during test time. This is a big problem for robust real world applications. 
The sensitive attribute relates now to the different domains or environments~\cite{Buhlmann2018}. As seen in Sec.~\ref{sec:initialExp} we enforce to learn a more general feature representation and to ignore domain specific attributes. This is leveraged to learn representations which are generic across (related) domains and hence would generalize better.




\paragraph{From MNIST to MNIST-R.}
Additionally to the \emph{MNIST} dataset we created the \emph{MNIST-R} dataset containing all original MNIST images rotated by 30 degrees. We train a simple neural net with two 128- and one 20-width ReLU representations. MNIST is used as source while the performance is measured on MNIST-R (target). Inspired by Heinze-Deml et al.~\cite{Heinze-Deml2017} few samples of the target set are used to improve the performance. We compare the results in Tab.~\ref{fig:MNIST-MNIST-R-DomainAdap} with a baseline model trained on the same amount of samples (20 per class) of the target dataset using data augmentation. Our approach can keep up with data augmentation, respectively even performs better if the imbalance in the amount of samples used during training becomes larger. It can better leverage the structure in the source data and map it to the target domain than simple data augmentation which relies on predefined transformations. 
\begin{SCfigure}[1][tb]%
   \centering
\includegraphics[width=0.5\textwidth]{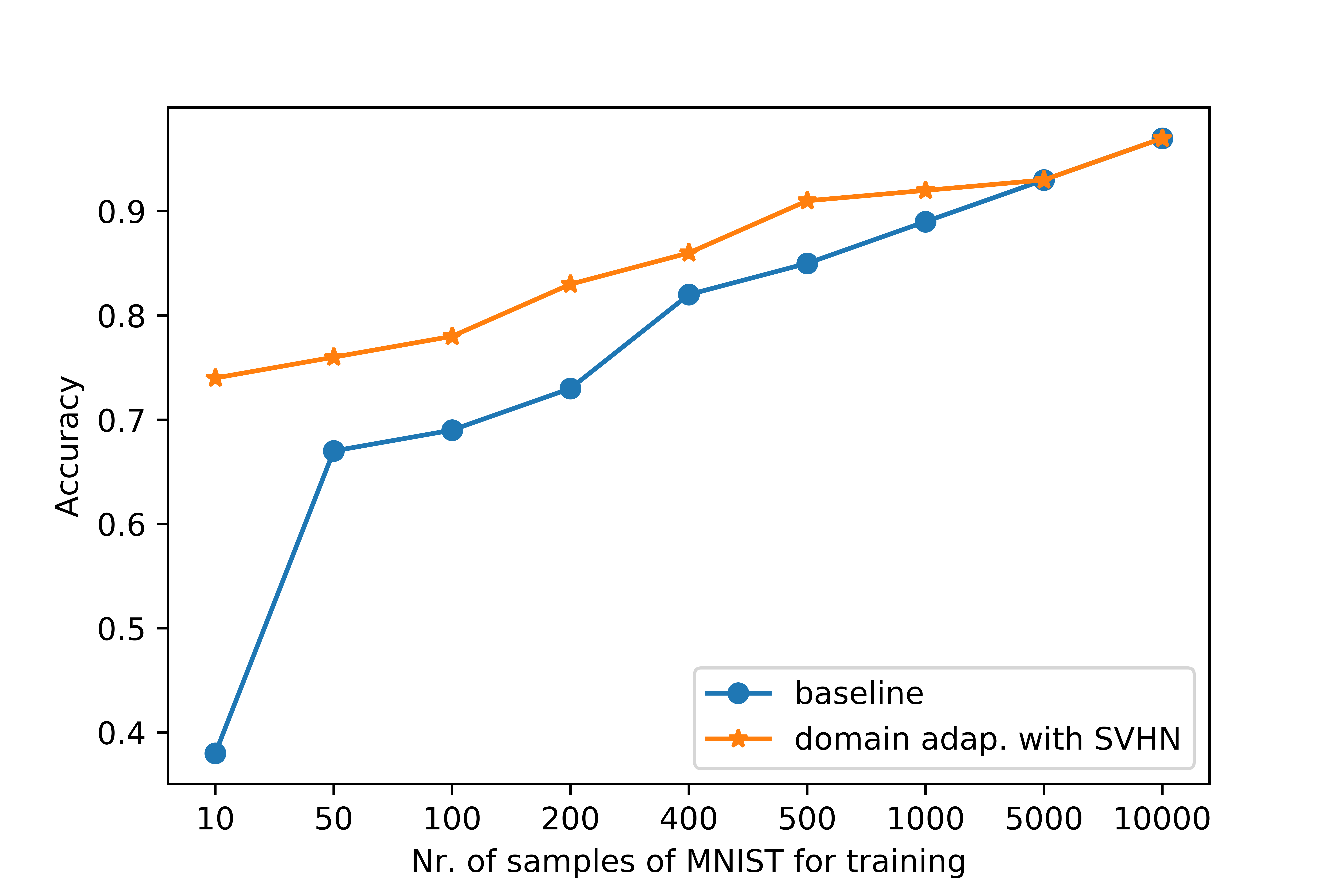}
\caption{Accuracy of MNIST with different number of training samples from the target domain (MNIST-R). Our approach (orange) clearly outperforms the simple baseline, especially when only few data from the target domain is provided.}
   \label{fig:MNISTvsMNISTSVHN}
\end{SCfigure} 

\paragraph{From SVHN to MNIST.}
The Street-View House Number (SVHN) dataset\footnote{\url{http://ufldl.stanford.edu/housenumbers/}, 2020/07/10.}, contains house numbers from Google Street View. The challenge of the SVHN dataset is the structured clutter in the background of images. A Convolutional Neural Network (CNN) with two double-Convolutional layers containing 32 and 64 nodes, respectively, is used. A 20-dim. feature representation on top of this architecture is applied to calculate the affinity loss. Results and comparison to Ganin et al.~\cite{Ganin2017} are shown in Tab.~\ref{tab:domainAdap}. The affinity loss does indeed improve the performance on the target dataset with only a little amount of samples. The performance of our model trained on the SVHN dataset with 10 MNIST samples reaches an accuracy of around 75\% and can be compared with Ganin et al.~\cite{Ganin2017}. In comparison, the baseline model achieves only an accuracy of 3\% (see Fig.~\ref{fig:MNISTvsMNISTSVHN}) on the same data. If there are only a few MNIST samples available, the neural net trained with SVHN and our affinity loss outperforms the baseline model trained solely on the same amount of MNIST samples. 

\begin{table}[tb]
  \centering
      \caption{Domain adaptation on MNIST (source) and MNIST-R (target). Displayed are the accuracies on the MNIST-R dataset after training (training dataset indicated in the header).}
    \begin{tabular}{|C{2cm}|C{2.3cm}C{3.5cm}C{3.9cm}|}
    \hline
    \textbf{\# source samples} &
    \textbf{only source dataset}& 
    \textbf{+ 200 target samples and data aug.} &
    \textbf{our approach (+ 200 target samples)}\\ \hline\hline
    
    1,000 & 0.52 & 0.76 & 0.78  \\
    10,000 & 0.67 & 0.76 & 0.82 \\
    \hline
    \end{tabular}
 %
    \label{fig:MNIST-MNIST-R-DomainAdap}
\end{table}%

\begin{table}[tb]
      \caption{Performance of models trained with the affinity loss on SVHN (source) and MNIST (target). For training a small amount of labeled target samples is used. Ganin et al.~\cite{Ganin2017} uses only unlabeled target samples for training.}
      \begin{tabular}{|l|l|C{3.5cm}C{3.5cm}|}
    \hline
     & & \textbf{accuracy SVHN (source domain)} & \textbf{accuracy MNIST (target domain)}\\\hline\hline
    & only SVHN & 0.91 & 0.11 \\\hline
    \multirow{2}{1.5cm}{baseline}&+ 100 MNIST samples & 0.91 & 0.11 \\
    & + 200 MNIST samples & 0.91 & 0.13 \\\hline
    \multirow{3}{1.5cm}{our approach}&+ 10 MNIST samples & 0.91 & 0.75 \\
    & + 100 MNIST samples & 0.91 & 0.80 \\
    & + 200 MNIST samples & 0.92 & 0.85 \\\hline
    &Ganin et al.~\cite{Ganin2017} & -  &0.74\\ \hline 
    \end{tabular}
    \label{tab:domainAdap}
\end{table}

%% file: discussion.tex
    We proposed a new approach for learning invariant feature representation. The main idea is to bring the feature representation of different distributions closer together by introducing an additional loss. We applied this strategy to three different areas: fairness, interpretability and domain adaptation. Our proposed method can be used for different model architectures as well as for readjusting the feature representation of existing, already trained models. 
    Experiments show that the equality gap can be significantly reduced while the accuracy is still kept at an acceptable level. The results are comparable with state-of-the-art methods for each task. We demonstrate how to understand how a sensitive attribute influences the classification of an input sample. 
    A challenge in our approach is to efficiently find the nearest neighbors in the embedding space. We rely on effective, approximated methods here. Not much thematized in this paper is that our approach allows using multiple source and target datasets. Thus a model can be trained to be fair regarding multiple attributes. A further extension might be using real-valued attributes as sensitive attributes.

%% file: main.bbl
\begin{thebibliography}{10}
\providecommand{\url}[1]{\texttt{#1}}
\providecommand{\urlprefix}{URL }

\bibitem{Adel}
Adel, T., Valera, I., Ghahramani, Z., Weller, A.: One-network adversarial
  fairness. In: AAAI Conference on Artificial Intelligence. AAAI (2019)

\bibitem{Amsterdam2019}
van Amsterdam, W.A.C., Verhoeff, J.J.C., de~Jong, P.A., Leiner, T., Eijkemans,
  M.J.C.: {Eliminating biasing signals in lung cancer images for prognosis
  predictions with deep learning}. npj Digital Medicine  2 (2019)

\bibitem{Beutel}
Beutel, A., Chen, J., Zhao, Z., Chi, E.H.: {Data Decisions and Theoretical
  Implications when Adversarially Learning Fair Representations}. Tech. rep.,
  Google Research (2017)

\bibitem{Buhlmann2018}
B{\"{u}}hlmann, P.: {Invariance, Causality and Robustness}. Tech. rep., ETH
  Zurich (2018)

\bibitem{Chattopadhyay2019}
Chattopadhyay, A., Manupriya, P., Sarkar, A., Balasubramanian, V.N.: {Neural
  Network Attributions: A Causal Perspective}. In: International Conference on
  Machine Learning (2019)

\bibitem{Ganin2017}
Ganin, Y., Ustinova, E., Ajakan, H., Germain, P., Larochelle, H., Laviolette,
  F., Marchand, M., Lempitsky, V.: {Domain-adversarial training of neural
  networks}. Journal of Machine Learning Research  17 (2017)

\bibitem{Ghifary}
Ghifary, M., Kleijn, W.B., Zhang, M., Balduzzi, D.: {Domain Generalization for
  Object Recognition with Multi-task Autoencoders}. In: International
  Conference on Computer Vision (2015)

\bibitem{HardtGoogle}
Hardt, M., Price, E.P., Srebro, N.: Equality of opportunity in supervised
  learning. In: Conference on Neural Information Processing Systems (2016)

\bibitem{Hartford2016}
Hartford, J., Lewis, G., Leyton-Brown, K., Taddy, M.: {Counterfactual
  Prediction with Deep Instrumental Variables Networks}. Tech. rep., Microsoft
  Research and University of British Columbia (2016)

\bibitem{Heinze-Deml2017}
Heinze-Deml, C., Meinshausen, N.: {Conditional Variance Penalties and Domain
  Shift Robustness}. Tech. rep., ETH Zurich (2017)

\bibitem{Huang}
Huang, J., Smola, A., Gretton, A., Borgwardt, K., Sch{\"{o}}lkopf, B.:
  Correcting sample selection bias by unlabeled data. In: Conference on Neural
  Information Processing Systems (2006)

\bibitem{Louizos2017}
Louizos, C., Shalit, U., Mooij, J., Sontag, D., Zemel, R., Welling, M.: Causal
  effect inference with deep latent-variable models. In: Conference on Neural
  Information Processing Systems (2017)

\bibitem{Louizos}
Louizos, C., Swersky, K., Li, Y., Welling, M., Zemel, R.: {The Variational Fair
  Autoencoder}. In: International Conference on Learning Representations (2015)

\bibitem{Magliacane2017}
Magliacane, S., van Ommen, T., Claassen, T., Bongers, S., Versteeg, P., Mooij,
  J.M.: {Domain Adaptation by Using Causal Inference to Predict Invariant
  Conditional Distributions}. In: Advances in Neural Information Processing
  Systems (2017)

\bibitem{Mancini2018}
Mancini, M., Porzi, L., Bul{\`{o}}, S.R., Caputo, B., Ricci, E.: {Boosting
  Domain Adaptation by Discovering Latent Domains}. In: Conference on Computer
  Vision and Pattern Recognition (2018)

\bibitem{Motiian2017}
Motiian, S., Piccirilli, M., Adjeroh, D.A., Doretto, G.: {Unified Deep
  Supervised Domain Adaptation and Generalization}. In: International
  Conference on Computer Vision (2017)

\bibitem{Pedreschi2008}
Pedreschi, D., Ruggieri, S., Turini, F.: {Discrimination-aware Data Mining}.
  In: ACM SIGKDD International Conference on Knowledge Discovery and Data
  Mining (2008)

\bibitem{Peters}
Peters, J., Janzing, D., Sch{\"{o}}lkopf, B.: {Elements of Causal Inference:
  Foundations and Learning Algorithms}. MIT Press, Cambridge, MA (2017)

\bibitem{Pinheiro2018}
Pinheiro, P.O.: {Unsupervised Domain Adaptation with Similarity Learning}. In:
  Conference on Computer Vision and Pattern Recognition (2018)

\bibitem{tinyImages}
Prabhu, V., Birhane, A.: {Large image datasets: A pyrrhic win for computer
  vision?} Tech. rep., UnifyID Inc. (2020)

\bibitem{Quadriantoa}
Quadrianto, N., Sharmanska, V.: Recycling privileged learning and distribution
  matching for fairness. In: Conference on Neural Information Processing
  Systems (2017)

\bibitem{Quadrianto}
Quadrianto, N., Sharmanska, V., Thomas, O.: {Discovering Fair Representations
  in the Data Domain}. In: Conference on Computer Vision and Pattern
  Recognition (2018)

\bibitem{Schumann2019}
Schumann, C., Wang, X., Beutel, A., Chen, J., Qian, H., Chi, E.H.: {Transfer of
  Machine Learning Fairness across Domains}. Tech. rep., Google (2019)

\bibitem{Singh2019}
Singh, H., Singh, R., Mhasawade, V., Chunara, R.: {Fair Predictors under
  Distribution Shift}. In: Conference on Neural Information Processing Systems
  (2019)

\bibitem{ThanhLuong2011}
{Thanh Luong}, B., Ruggieri, S., Turini, F.: {k-NN as an Implementation of
  Situation Testing for Discrimination Discovery and Prevention}. In: ACM
  SIGKDD International Conference on Knowledge Discovery and Data Mining (2011)

\bibitem{Verma2018}
Verma, S., Rubin, J.: {Fairness Definitions Explained}. In: IEEE/ACM
  International Workshop on Software Fairness (2018)

\bibitem{Xie}
Xie, Q., Dai, Z., Du, Y., Hovy, E., Neubig, G.: {Controllable Invariance
  through Adversarial Feature Learning}. In: Conference on Neural Information
  Processing Systems (2017)

\bibitem{Yang}
Yang, J., Yang, R., Hauptmann, A.G.: Adapting svm classifiers to data with
  shifted distributions. In: International Conference on Data Mining Workshops
  (2007)

\bibitem{Zafar2016}
Zafar, M.B., Valera, I., Rodriguez, M.G., Gummadi, K.P.: {Fairness Beyond
  Disparate Treatment {\&} Disparate Impact: Learning Classification without
  Disparate Mistreatment}. In: International World Wide Web Conference (2017)

\bibitem{Zemel2013}
Zemel, R., Ledell, Y.., Wu, ., Swersky, K., Pitassi, T., Dwork, C.: Learning
  fair representations. In: International Conference on Machine Learning.
  vol.~3. ICML (2013)

\end{thebibliography}
